%% file: main.tex
\documentclass[sigconf]{acmart}

\makeatletter
\def\@ACM@checkaffil{
    \if@ACM@instpresent\else
    \ClassWarningNoLine{\@classname}{No institution present for an affiliation}%
    \fi
    \if@ACM@citypresent\else
    \ClassWarningNoLine{\@classname}{No city present for an affiliation}%
    \fi
    \if@ACM@countrypresent\else
        \ClassWarningNoLine{\@classname}{No country present for an affiliation}%
    \fi
}
\makeatother

\usepackage{color}
\usepackage{multirow}
\usepackage{enumitem}
\setlist[enumerate]{leftmargin=*}

\AtBeginDocument{%
  \providecommand\BibTeX{{%
    \normalfont B\kern-0.5em{\scshape i\kern-0.25em b}\kern-0.8em\TeX}}}

\copyrightyear{2023} 
\acmYear{2023} 
\setcopyright{acmlicensed}\acmConference[MM '23]{Proceedings of the 31st ACM International Conference on Multimedia}{October 29-November 3, 2023}{Ottawa, ON, Canada}
\acmBooktitle{Proceedings of the 31st ACM International Conference on Multimedia (MM '23), October 29-November 3, 2023, Ottawa, ON, Canada}
\acmPrice{15.00}
\acmDOI{10.1145/3581783.3612595}
\acmISBN{979-8-4007-0108-5/23/10}

\acmConference[ACM MM '23]{the 31th ACM International Conference on Multimedia}{October 29--November 3, 2023}{Ottawa, Canada}

\acmSubmissionID{4151}
\begin{document}

\title{Learning Profitable NFT Image Diffusions via\\ 
Multiple Visual-Policy Guided Reinforcement Learning}

\author{Huiguo He}
\email{hehg3@mail2.sysu.edu.cn}
\affiliation{
  \institution{Sun Yat-Sun University}
  \country{}
}

\author{Tianfu Wang}
\email{tianfuwang@mail.ustc.edu.cn}
\affiliation{
  \institution{University of Science and Technology of China}
  \country{}
}

\author{Huan yang}
\authornote{
Corresponding authors: Huan Yang, Jianlong Fu, and Jian Yin.
}
\email{huayan@microsoft.com}
\affiliation{
  \institution{Microsoft Research Asia}
  \country{}
}

\author{Jianlong Fu}
\authornotemark[1]
\email{jianf@microsoft.com}
\affiliation{
  \institution{Microsoft Research Asia}
  \country{}
}

\author{Nicholas Jing Yuan}
\email{nicholas.yuan@microsoft.com}
\affiliation{
  \institution{Microsoft}
  \country{}
}

\author{Jian Yin}
\authornotemark[1]
\email{issjyin@mail.sysu.edu.cn}
\affiliation{
  \institution{Sun Yat-Sun University}
  \country{}
}

\author{Hongyang Chao}
\email{isschhy@mail.sysu.edu.cn}
\affiliation{
  \institution{Sun Yat-Sun University}
  \country{}
}

\author{Qi Zhang}
\email{zhang.qi@microsoft.com}
\affiliation{
  \institution{Microsoft}
  \country{}
}

\renewcommand{\shortauthors}{Huiguo He et al.} 

\begin{teaserfigure}
    \centering
    \vspace{-1.7em}
  \includegraphics[width=0.91\textwidth]{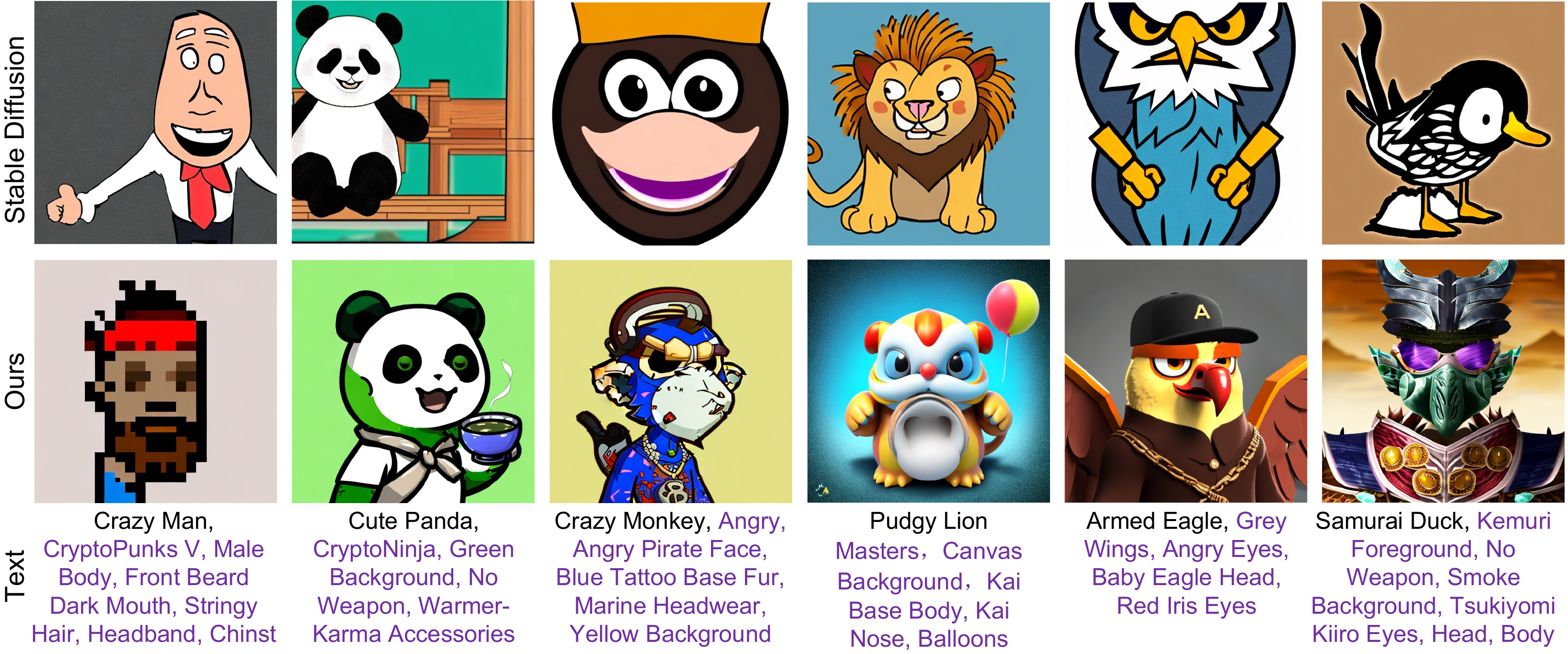}
    \vspace{-0.5cm}
  \caption{Comparisons between our approach and base Stable-Diffusion over six NFT categories, including \textit{pixel art, clip art, illustration, pseudo 3D, 3D, and complex patterns} from left to right. Compared to baselines, our approach creates more NFT-style images with more fancy decorations and visual experiences. The purple texts are completed properties, given user inputs.}
  \Description{Enjoying the baseball game from the third-base
  seats. Ichiro Suzuki preparing to bat.}
  \label{fig:teaser}
\end{teaserfigure}

\begin{abstract}
 We study the task of generating profitable Non-Fungible Token (NFT) images from user-input texts. Recent advances in diffusion models have shown great potential for image generation. 
 However, existing works can fall short in generating visually-pleasing and highly-profitable NFT images, mainly due to the lack of 1) plentiful and fine-grained visual attribute prompts for an NFT image, and 2) effective optimization metrics for generating high-quality NFT images. To solve these challenges, we propose a \textbf{Diffusion}-based generation framework with \textbf{M}ultiple \textbf{V}isual-\textbf{P}olicies as rewards (i.e., \textbf{Diffusion-MVP}) for NFT images. The proposed framework consists of a large language model (LLM), a diffusion-based image generator, and a series of visual rewards by design. First, the LLM enhances a basic human input (such as ``panda'') by generating more comprehensive NFT-style prompts that include specific visual attributes, such as ``panda with Ninja style and green background.'' Second, the diffusion-based image generator is fine-tuned using a large-scale NFT dataset to capture fine-grained image styles and accessory compositions of popular NFT elements. Third, we further propose to utilize multiple visual-policies as optimization goals, including visual rarity levels, visual aesthetic scores, and CLIP-based text-image relevances. This design ensures that our proposed \textbf{Diffusion-MVP} is capable of minting NFT images with high visual quality and market value. To facilitate this research, we have collected the largest publicly available NFT image dataset to date, consisting of 1.5 million high-quality images with corresponding texts and market values.  Extensive experiments including objective evaluations and user studies demonstrate that our framework can generate NFT images showing more visually engaging elements and higher market value, compared with state-of-the-art approaches. 
 \end{abstract}



\maketitle

\section{Introduction}

Creating Non-Fungible Token (NFT) images has gained tremendous popularity in recent years, because of their visual uniqueness, attractiveness, and richness of various gorgeous elements. The digital ownership of NFT images has revolutionized the art world, and opened up new avenues for the creation and sale of these unique digital assets that can be bought, sold, and traded like physical artwork. It is reported that the NFT market is expected to significantly grow at an annual rate of 35.0\%, reaching \$13.6 billion by 2027\footnote{\href{https://www.marketsandmarkets.com/Market-Reports/non-fungible-tokens-market-254783418.html?gclid=Cj0KCQjw2v-gBhC1ARIsAOQdKY1GJG52B45LCMLza6vDu6YIhgIvK-EKArG8AAb5EKMhS_Gle_RFATAaAh8rEALw_wcB}{marketsandmarkets.com}}. Despite of the popularity of NFT images, the design of such art-style images is still a challenging task. The art world has always been creating something that is unique, and visually appealing. To achieve this goal, human artists need to consider various factors such as the aesthetic appeal, rarity, and uniqueness of their creations to ensure that they can stand out in the competitive NFT market. Moreover, as the demand for NFT images continues to grow, there is a need for innovative and distinctive designs to capture the attention of real markets that usually reflects the demand from potential buyers and collectors.

With the emergence of Artificial Intelligence Generated Content (AIGC), in this paper, we take one step further to study the possibility of generating profitable NFT images. Recently, in image generation domains, great results have been created using Variational AutoEncoders (VAEs)~\cite{VAE}, Generative Adversarial Networks 
(GAN)~\cite{karras2020impro_stylegan, reed2016generative_T2I, dong2017semanticGAN2img, zhang2017stackgan_T2I, zhang2018stackgan++_T2I, xu2018attngan_T2I, zhang2018photographic_T2I, qiao2019mirrorgan_T2I,  zhu2019dm_T2I,tao2020dfGAN_T2I}, and Diffusion-based models~\cite{song2020DDIM_diffusion, DDPM_diffusion, song2020score_diffusion, nichol2021improved_diffusion, dhariwal2021_diffusion, saharia2022palette, lu2022DPM,nichol2021CLIDE_diffusion, ramesh2021zero_T2I_diffusion, saharia2022photorealistic, ramesh2022hierarchical_DALLE2, sd_ldm_diffusion}. 
The most recent success has been made by Stable Diffusion~\cite{sd_ldm_diffusion}, which achieves state-of-the-art results by performing diffusion in a latent space to reduce computational cost while maintaining excellent visual performance. Due to its ease of use and naturalness, Stable-Diffusion has been widely used in a variety of text-to-image applications ~\cite{ruiz2022dreambooth, lhhuang2023composer}. 

\begin{figure}[t]
\setlength{\abovecaptionskip}{-0.05cm}
\centering
\vspace{-10pt}
\includegraphics[width=0.95\columnwidth]{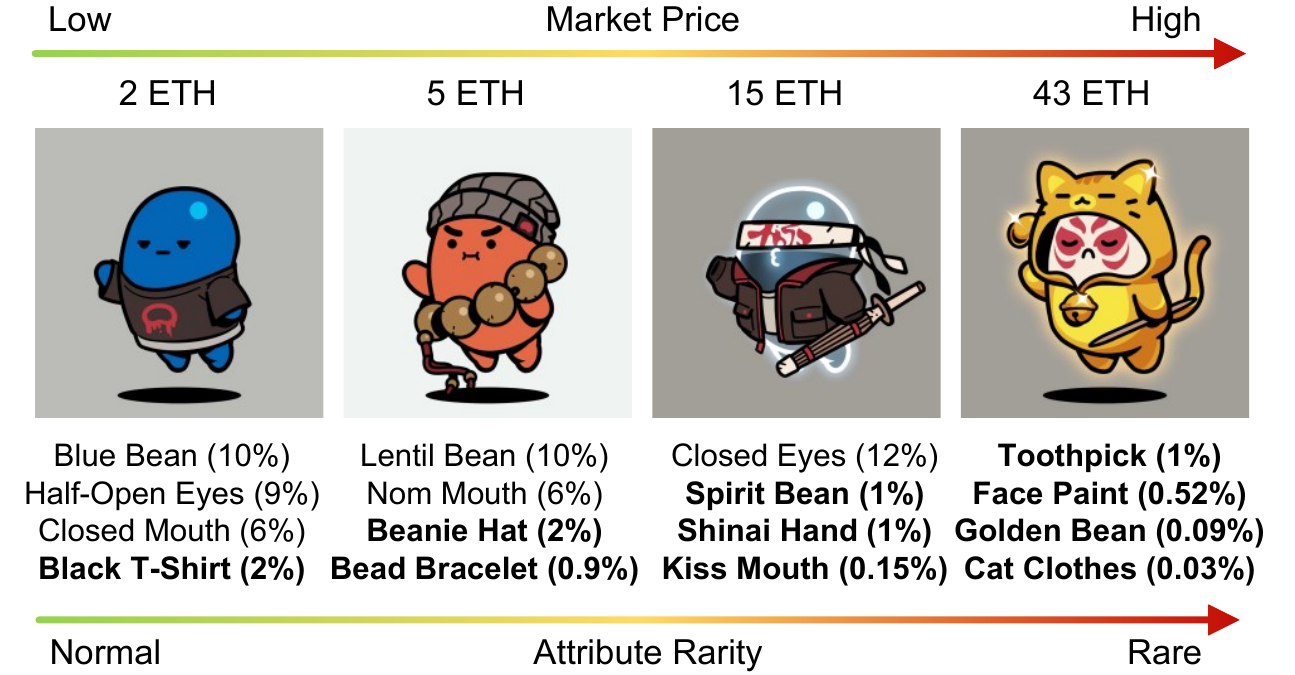}
\caption{Four NFT examples in the BEANZ collection. The texts indicate corresponding attributes with rarity scores (the lower, the rare). The prices are represented by ETH, and the peak Ethereum price is about $2.1k$ USD in April 2023. The more rare of the attributes, the higher price of NFT images.}
\label{fig:NFT_display}
\vspace{-20pt}
\end{figure}

Although promising visual results have been created, there are still grand challenges for current image generation models to create visually appealing and profitable NFT images. The reason lies in two folds. First, existing models are mainly trained by general image datasets (e.g., LAION-5B~\cite{schuhmann2022laion_5B}), usually lacking of fine-grained NFT-type attribute descriptions that play a key role in generating fancy and profitable images. For example in Fig.~\ref{fig:NFT_display}, there are rich descriptions for the NFT images, such as ``golden'', ``cat clothes''. Without these detailed attribute descriptions, NFT images can only show limited creativities, and thus lead to poor market values. Second, existing models are often short of suitable optimization metrics in training, which is difficult for models to generate popular characteristics that meet collectors' preferences in the market. Note that supervised-training by pixel-wise losses (e.g., in SD~\cite{sd_ldm_diffusion}) on NFT image datasets can only help to learn NFT visual styles. However, how to generate highly profitable NFTs with rarely visual attributes is still largely under-explored.

To address the above issues, we propose a novel image \textbf{Diffusion} model for NFT image generation by optimizing \textbf{M}utliple \textbf{V}isual-\textbf{P}olicies (denoted as \textbf{Diffusion-MVP}). Specifically, given a user input (e.g., ``panda'') for an NFT topic, \textbf{Diffusion-MVP} first utilizes a large language model (LLM) like GPT-2 to complete the user input by generating plentiful NFT attributes for the object ``panda''. To generate such rich attributes, the LLM is fine-tuned on large-scale NFT image descriptions by randomly masking out attribute terms, and predicting them from objects in turn. Second, we propose to adopt the Stable-Diffusion model as our base image generator, and fine-tune the model on a NFT image dataset to acquire the NFT image styles and accessory compositions of popular NFT elements. Third, to generate NFT images with higher market values, we propose to utilize multiple visual-policies that optimize the base image generator by reinforcement learning. Such a design ensures to generate NFT images equipped with visually pleasant and rare elements mining from real markets, and thus can significantly increase the market value of generated NFT images. In particular, we propose to design a visual rarity classifier, and adopt a visual aesthetic scoring model, and a CLIP-based text-image relevance model, as a combination of visual policies in training.

To facilitate this research, we have collected and published to-date the largest NFT image datasets, which consists of \textbf{1.5 million} high-resolution images with corresponding texts and real market value. Extensive experiments demonstrate the effectiveness of the proposed \textbf{Diffusion-MVP} compared with several competitive baselines including base Stable-Diffusion and DALL$\cdot$E\ 2 models, by using both objective and subjective evaluation metrics. A user study with over 2k votes from 10 human subjects further shows dominant preferences to our approach. To better promote the research for NFT image generation, we will release both datasets and models in the future.

\section{Related Works}

\begin{figure}[t]
    \setlength{\abovecaptionskip}{-0.05cm}
    \centering
    \includegraphics[width=0.95\columnwidth]{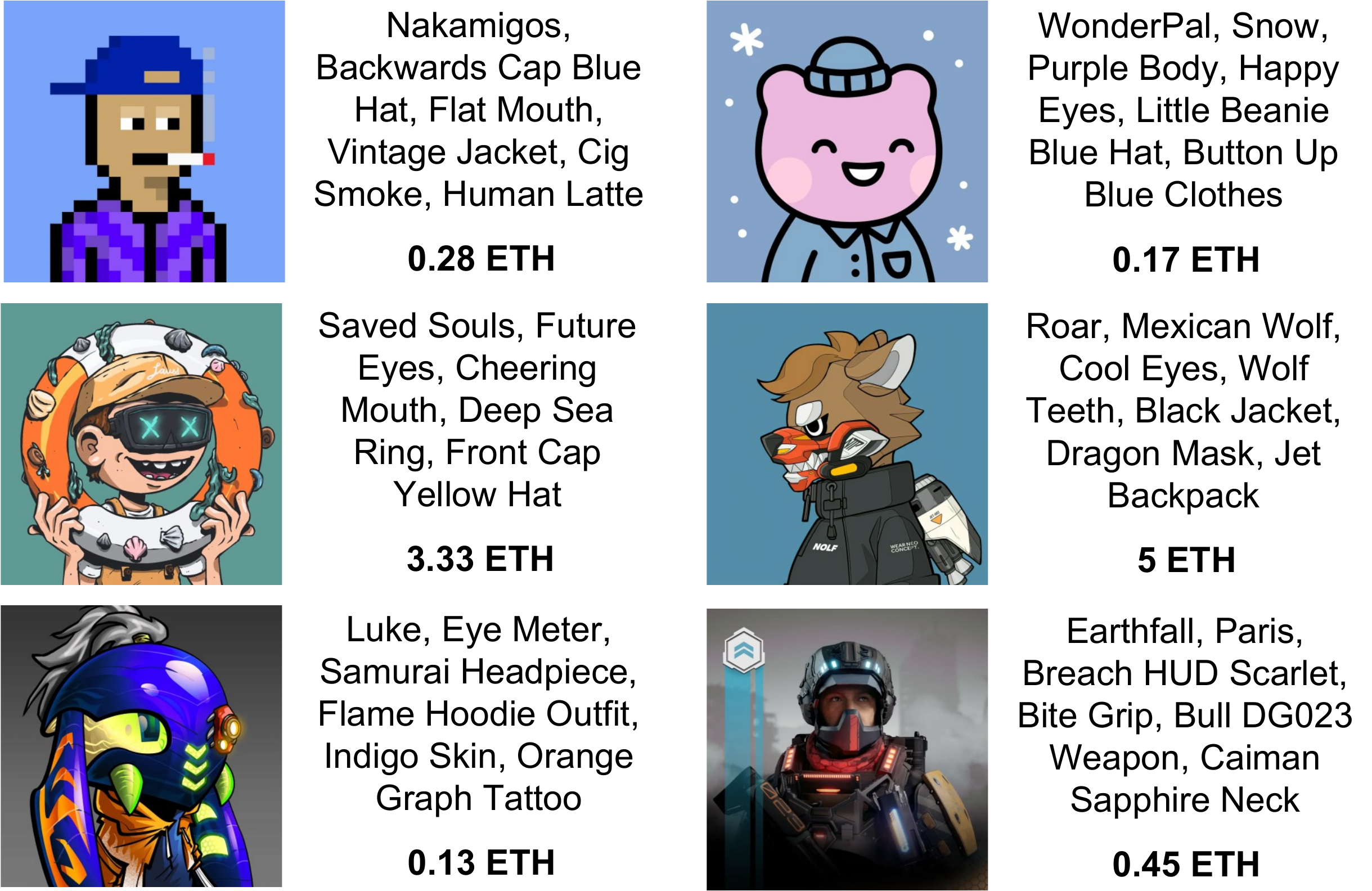}
    \caption{Engaging NFT examples with property descriptions, and ETH prices, from our newly-collected \textbf{NFT-1.5M} dataset. }
    \label{fig:data_Statis}
    \vspace{-0.4cm}
\end{figure}

\subsection{Image Generation}
\ 

Image generation has consistently been a popular research topic within the field of computer vision. Early research primarily focused on Variational Autoencoders (VAEs)~\cite{VAE}, flow-based methods~\cite{kingma2013flow1, dinh2016flow2}, Generative Adversarial Networks (GANs)~\cite{brock2018largeGAN, karras2020impro_stylegan, reed2016generative_T2I, dong2017semanticGAN2img, zhang2017stackgan_T2I, zhang2018stackgan++_T2I, xu2018attngan_T2I, zhang2018photographic_T2I, qiao2019mirrorgan_T2I,  zhu2019dm_T2I,tao2020dfGAN_T2I, tiankai2023language, ma2022ai}. While the sampling quality of VAEs and flow-based methods is inferior to that of GANs~\cite{GAN}, GANs have optimization difficulties~\cite{arjovsky2017WGAN, gulrajani2017improvedWGAN, metz2016unrolledGAN} that limit their performance. 
In addition, some researchers have attempted to improve the quality of generated images by Super-Resolution (SR)~\cite{wang2021realesrgan, yang2020learning, liu2022learning, liu2022ttvfi, yang2022degradation, qiu2022learning, zixi2023VQGAN, chan2022basicvsr++, liang2021swinir} or by increasing diversity~\cite{SelfConditionedGANs, TediGAN, li2022talk2face}. However, these methods are not the key to increasing the value of NFTs.

Recently, diffusion-based generative models~\cite{song2020DDIM_diffusion, DDPM_diffusion, song2020score_diffusion, nichol2021improved_diffusion, dhariwal2021_diffusion, saharia2022palette, ruan2022mm_diffusion, lu2022DPM, ma2023unified, wang2023videofactory, zhu2023moviefactory} have emerged, achieving state-of-the-art results in terms of image quality and diversity. As a result, diffusion-based text-to-image 
generation~\cite{nichol2021CLIDE_diffusion, ramesh2021zero_T2I_diffusion, ramesh2022hierarchical_DALLE2, sd_ldm_diffusion, ruiz2022dreambooth, saharia2022photorealistic, lhhuang2023composer} received much attention from academia and industry due to the simplicity and naturalness of text control. Specifically, Stable Diffusion (SD)~\cite{sd_ldm_diffusion} applies the diffusion model to the latent space and is currently the SOTA open-source image generation model trained on LAION-5B~\cite{schuhmann2022laion_5B}, the largest general image-text pair dataset. However, existing generative models are trained on general images and lack domain knowledge of NFT, which will lead to suboptimal performances in NFT image generation.

\subsection{Reinforcement Learning}
\ 
Reinforcement Learning (RL) seeks to maximize cumulative rewards received by an agent through its interactions with an environment.
Many works, including value-based approaches~\cite{DQN} and actor-critic approaches~\cite{A3C, TRPO, ppo}, have been proposed to solve this optimization problem. Among them, PPO~\cite{ppo} is a popular actor-critic approach where an actor selects actions and a critic evaluates the decision quality.
PPO adopted a clipped surrogate objective function, achieving safer and more stable optimization than A3C~\cite{A3C} while simplifying the complexity compared to TRPO~\cite{TRPO}.

Recently, several studies have attempted to fine-tune large language models using Reinforcement Learning from Human Feedback (RLHF)~\cite{NIPS2020sumHumanFeedback, deepmind2022teaching, ouyang2022humanFeedback, lu2022quark, reinforcement2022ICLR23} and AI feedback~\cite{AIfeedback2022constitutional}.
For instance, Stiennon et al.~\cite{NIPS2020sumHumanFeedback} trained language models to improve summarization using human feedback. Menick et al.~\cite{deepmind2022teaching} used RLHF to train an "open book" question-answering model that generates answers while citing specific evidence to support its claims. Quantized Reward Konditioning (Quark)~\cite{lu2022quark} is proposed for optimizing a reward function that quantifies (un)wanted properties. Ouyang et al.~\cite{ouyang2022humanFeedback} proposed InstructGPT, which fine-tunes language models to better follow user intent using human feedback. Ramamurthy et al.~\cite{reinforcement2022ICLR23} proposed Natural Language Policy Optimization (NLPO) to effectively reduce the combinatorial action space in language generation. 
While existing methods have shown that reinforcement learning can improve model performance through various rewards, few have incorporated value information from the NFT market. This paper aims to mine value information from the NFT market and incorporate it into the image generation process to enhance the value of generated NFT images.

\section{NFT-1.5M Dataset}\label{sec:dataset}

\begin{table}[t]
\setlength{\abovecaptionskip}{-0.05cm}
\setlength{\belowcaptionskip}{-0.2cm} 
\centering
\caption{Data statistics of \textbf{NFT-1.5M} dataset, over image quality, textual property, and price range. We use Ethereum (ETH) price for value range.}
\label{tab:dataset}
\begin{tabular}{cc|cc|cc}
\hline
\multicolumn{2}{c|}{\textbf{Image Resolution}} & \multicolumn{2}{c|}{\textbf{Property Number}} & \multicolumn{2}{c}{\textbf{Price Range}} \\ \hline
Value & Ratio & Value & Ratio & Value & Ratio \\ \hline
$\leq$ 1K & 25\% & $\leq$ 6 & 45\% & $\leq$ 1 & 65\% \\
1K $\sim$ 2K & 45\% & 6 $\sim$ 10 & 43\% & 1 $\sim$ 10 & 30\% \\
\textgreater 2K & 30\% & \textgreater 10 & 12\% & \textgreater 10 & 5\% \\ \hline
\end{tabular}
\end{table}

In this section, we present the construction of our newly-collected NFT dataset. We will first introduce the full dataset \textbf{NFT-4M}, which consists of 4 million text-image-value triplet pairs. \textbf{NFT-4M} comprises the top 1,000 collections with the highest total transaction value. All NFT information was obtained by crawling OpenSea\footnote{\href{https://opensea.io/}{opensea.io}}, which is the largest NFT market website. The dataset can be used for multiple tasks, such as NFT generation, price prediction, etc. In this paper, we mainly use the dataset to mine the potential relationship between NFT values and visual features for profitable NFT image generation. In Sec.~\ref{sec:market_value}, we first introduce our rarity score definition, which is highly related to NFT value while eliminating the noise of infrequent NFT trading and WEB3 market fluctuation. Later in Sec.~\ref{sec:clean_data}, to fine-tune our NFT image generator with high visual quality, we further cleaned the dataset to 1.5 million (called \textbf{NFT-1.5M} subset).

\subsection{NFT Image Pricing} \label{sec:market_value} 
\ 

The price of NFT images can vary a lot. An intuitive approach to value an NFT image is to use its current market price. However, NFT market transactions are infrequent, often resulting in potentially non-existent or lagging transaction prices, bids, and asks. Additionally, as the overall WEB3 market price fluctuates greatly, using such a lagging price to value NFTs may be inaccurate. Fortunately, studies have shown that the NFT rarity highly correlates with their prices~\cite{mekacher2022SCI}. Besides, third-party NFT valuation platforms like NFTBank\footnote{\href{https://nftbank.ai/}{nftbank.ai}}, Mintable\footnote{\href{https://mintable.app/}{mintable.app}}, and Rarible\footnote{\href{https://rarible.com/}{rarible.com}} also adopt NFT rarity as a key factor in valuation. Inspired by these works and platforms, we rank NFT rarity within a collection and define relative value according to this ranking. Here we define the value of an NFT by its rarity ranking within the collection, which is defined  as follows:
\begin{equation}
    V_r = \sum_{i \in \Omega} \frac{1}{\eta_i} ,
\end{equation}
where the $\eta_i$ represents the proportion of NFTs with the property $i$ in the entire collection and $\Omega$ represents all the properties of this NFT. Compared to defining NFT value using market price, this definition has the following three advantages: 1) it eliminates the impact of WEB3 market price fluctuations; 2) it overcomes the problem of noise in defining value using lagging NFT prices; 3) it ignores the influence between different collections and weakens the impact of community marketing on the value of NFTs. 

To predict the range of NFT image values, we divided NFT value ranges into three tiers followed by a recent study on NFT selling price prediction~\cite{NFTWWW2023show}. NFTs within a collection were ranked based on their rarity scores and categorized into three rarity levels accordingly. We defined the top 5\% ranges as \textbf{high-priced} NFTs. And the remaining categories were approximately equally divided into \textbf{medium-priced} (top 5\%-60\%) and \textbf{low-priced} (top 60\%-100\%). Based on the above price definition, each NFT image in the proposed \textbf{NFT-4M} datasets can have a reasonable valuation.

\subsection{Dataset Cleaning for Image Generation} \label{sec:clean_data}

As the NFT market has a variety of collections, the data inevitably contains undesirable data items, such as non-image data and meaningless text. It is necessary to further clean the collected dataset with better quality for NFT text-to-image generation tasks. Therefore, we conduct the following cleaning procedures step-by-step:

\begin{enumerate}
\item \textbf{Non-images filter:} about 10.17\% percentages of non-image data among the total dataset, such as MP4 and gif files were removed.
\item \textbf{Resolution filter:} about 14.81\% percentages of images with low resolutions or non-square shapes were removed.
\item \textbf{Property filter:} collections with less than 3 properties were removed due to their insufficient NFT text information.
\item \textbf{Visual content filter:} a visual content clustering algorithm~\cite{McInnes2017cluster} was utilized to filter out NFTs (e.g., the class of virtual world passports), whose visual content contains a significant number of metaverse URLs, addresses, or other irrelevant texts. 
\item \textbf{Duplicate filter:} collections with high intra-collection image similarity were removed by duplicate detection. 
\end{enumerate}
After the above cleaning steps, an NFT dataset comprising approximately 1.5 million image-text-value pairs was constructed and designated as \textbf{NFT-1.5M} subset for NFT image generation. It contains high-quality text-image data pairs, in which images are mainly designed by artists, and texts are carefully annotated by NFT creators. As a result, it is suitable for text-to-image generation training. Examplar NFT images from the \textbf{NFT-1.5M} dataset are shown in Fig.~\ref{fig:data_Statis}, and data statistics over image resolution, NFT properties, and prices are presented in Tab. ~\ref{tab:dataset}. 

\begin{figure*}[t]
\setlength{\abovecaptionskip}{-0.1cm} 
\setlength{\belowcaptionskip}{-0.4cm} 
\centering
\includegraphics[width=0.98\linewidth]{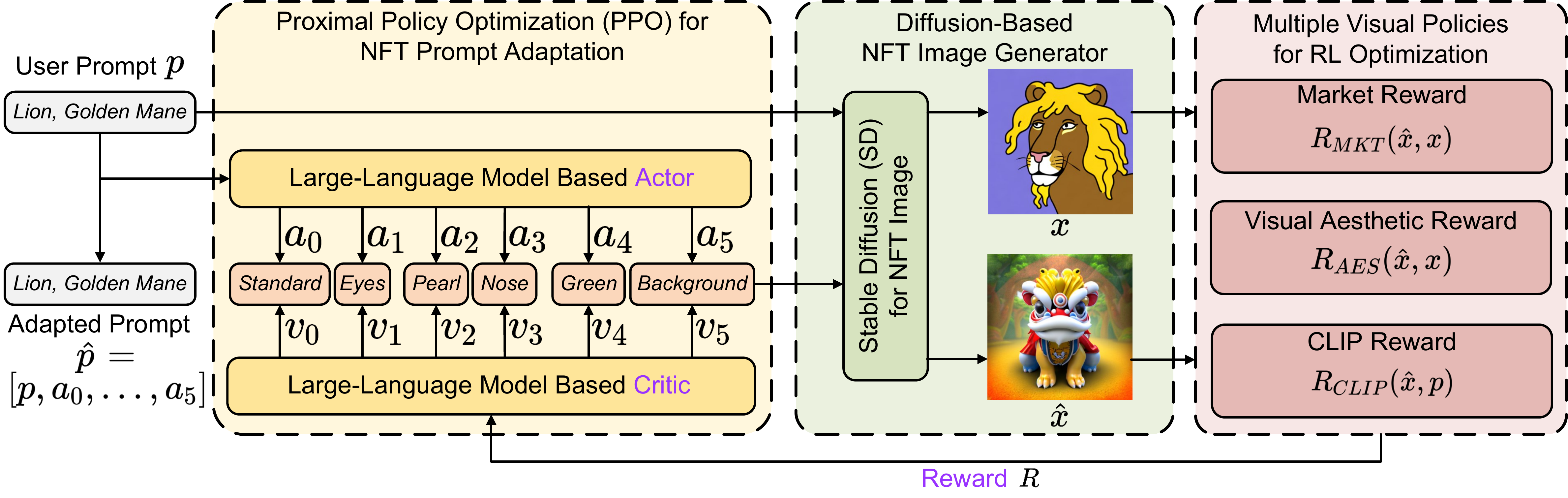}
\caption{Framework of \textbf{Diffusion-MVP} for text-based NFT image generation. The image generator in the middle takes enhanced NFT attribute prompts as input, which is further optimized by novel multiple visual-policies as combined rewards, to reflect market rarity, aesthetic score, and text-image relevance. Rewards are fed into Prompt Adaptation module as critic value (indicated by $v$), and actions (indicated by $a$) are conducted to improve NFT attribute prompts for image generators.}
\label{fig:framework}

\end{figure*}

\section{Our Approach}\label{sec:method}

In this section, we introduce our \textbf{Diffusion}-based generation framework with \textbf{M}ultiple \textbf{V}isual \textbf{P}olicies as rewards for NFT images (denoted as \textbf{Diffusion-MVP}). The framework of \textbf{Diffusion-MVP} is shown in Fig.~\ref{fig:framework}. First, when users input a prompt, e.g., "Lion, Golden Mane", an \textbf{RL-based NFT prompt adaption module} will rewrite the prompt by adding visually pleasing, rare, and popular fine-grained attribute descriptions, such as "Pearl Nose". Second, an \textbf{image generator} built on Stable-Diffusion for NFT images can capture these fine-grained descriptions to generate high-quality and profitable NFT images. Third, due to the optimized metrics are crucial for generating high-quality NFT images, we adopted three carefully-designed \textbf{visual policies as rewards} to guide the optimization direction of RL training. In the following sections, we introduce our three main modules step-by-step: 1) LLM with PPO framework in Sec.~\ref{sec:PPO}, 2) NFT Image generator in Sec.~\ref{sec:SD}, and 3) visual multi-reward in Sec.~\ref{sec:Reward}. And finally in Sec.~\ref{sec:training}, we will present the overall optimization target. 

\subsection{Optimizing Prompts with PPO Framework}\label{sec:PPO}

Recent studies have shown that designing suitable prompts are crucial for generating high-quality images in text-to-image 
methods~\cite{ramesh2021zero_T2I_diffusion,hao2022promptist_furu}.
In our framework, the LLM (i.e., the yellow module in Fig.~\ref{fig:framework}) modifies user input prompts to produce language that better fits the following NFT image generator. Its goal is to create more profitable NFT images by adding detailed, appealing, and popular elements to original prompts. To learn fine-grained descriptions in the NFT domain, we first adopt Supervised-FineTune (denoted as \textbf{SFT}) for the LLM model by using our \textbf{NFT-1.5M} dataset. By randomly masking some attributes, the LLM learns to predict these missing attribute descriptions. To further improve the performance of the LLM for generating better prompts, and tap into the potential value of the NFT market, we propose to adopt a reinforcement learning strategy to further improve the performance of the above SFT-LLM. The paradigm has been proven effective to enhance model generalization capability in previous reinforcement learning works~\cite{ouyang2022humanFeedback, NIPS2020sumHumanFeedback, AIfeedback2022constitutional, hao2022promptist_furu}.

Reinforcement Learning (RL) is a type of machine learning that aims to train an agent to make decisions by maximizing cumulative rewards. The agent continuously interacts with its environment by observing a state $s$, selecting an action $a$ based on its policy $\pi$, and then receiving a reward $R$. The policy $\pi$ of the agent is an actor parameterized by $\theta$. The probability of this actor taking a sequence of actions, or a trajectory $\tau$, is denoted as $\Pr(\tau | \pi)$. The optimization of the objective function is as follows:
\begin{equation}\label{eqn:RL_goal}
    J(\pi) = \int_\tau \Pr(\tau | \pi_{\theta})G (\tau),
\end{equation}
where the $G (\tau)$ is the return of the trajectory $\tau$ that can be obtained from the sum of the discounted reward. 
PPO is widely adopted to stabilize the RL training process due to its high performance and efficiency~\cite{ppo}. To explain clearly, it is an Actor-Critic method where the actor controls the agent’s behavior and the critic evaluates the quality of actions.

In this paper, we employ a pre-trained GPT-2 model followed by several adaptation layers of MLPs to serve as our actor. The critic has a similar architecture to the actor (except for the output dimension) and shares the same GPT-2 backbone with the actor. As shown in Fig.~\ref{fig:framework}, the LLM can be seen as the actor in our scenario, interacting with an imaginary NFT market (represented by the Market Reward), and receiving multiple visual policies as rewards. The LLM outputs a token from a pre-defined vocabulary, which is similar to how an actor chooses an action $a$ to perform from an action space. The sentence $p=\{p_0, a_0, a_1, \cdots \}$ output by the LLM, consisting of a string of tokens, corresponds to the trajectory $\tau$ in RL. The NFT image generator takes the adapted prompt generated by the LLM as input to create NFT images. The resulting images are then evaluated by multiple visual policies, which provide feedback and rewards to the LLM. Therefore, our LLM can be treated as an actor, which is optimized in a PPO manner. By interacting with this environment, our LLM has the opportunity to explore the trajectories (i.e., sentences in our scenario) unseen in the training dataset and can obtain further improvements compared to SFT-LLM. The policy gradient loss of LLM is presented as follows:
\begin{equation}\label{eqn:PPO_policy_loss_master}
    \mathcal{L}_\text{\scriptsize \tiny PG} = \min( \frac{\pi_\theta(a_t|s_t)}{\pi_{\theta_{\text{\scriptsize \tiny SFT}}}(a_t|s_t)} A(s_t, a_t), \,\, g(\epsilon, A(s_t, a_t))   ),
\end{equation}
\begin{equation}\label{eqn:PPO_policy_loss_2}
    g(\epsilon, A) = 
    \begin{cases}
    (1+\epsilon)A, & A \geq 0 \\
    (1-\epsilon)A, & A < 0
    \end{cases}
\end{equation}
where the clip function $g$ is adopted to enhance the stability of policy optimization. $A(s_t, a_t) = G_t - V_\phi(s_t)$ is the advantage function to measure the relative quality of token action compared to average quality. $G_t$ is the total discounted reward obtained from the timestep $t$ onwards. $V_\phi(s_t)$ is the predicted expected return of the state $s_t$ from a critic, which is optimized using MSE loss with respect to the critic's parameter $\phi$:
\begin{equation}\label{eqn:PPO_value_loss_3}
    \mathcal{L}_\text{\scriptsize \tiny V} = \mathbb{E}_{t, s}(V_\phi(s_t) - G_t)^2.
\end{equation}

\subsection{NFT Image Generator}\label{sec:SD}
\ 

Existing image generation approaches mainly train on general image datasets. This is sub-optimal for NFT image generation due to the lack of domain knowledge of NFTs. As shown in Fig.~\ref{fig:NFT_display}, currently popular NFTs are mainly anthropomorphic and contain rich fine-grained attributes, which differ significantly from general images. To generate high-quality NFT images, we fine-tuned the state-of-the-art image generation model, i.e., Stable Diffusion (SD)~\cite{sd_ldm_diffusion}, on our \textbf{NFT-1.5M} dataset to learn the style and characteristics of NFTs. In the following, we will elaborate on how we fine-tune the Stable Diffusion to better match NFT domains. 
 
Defining $x_0$ as a sample in the data $X$, the forward process in diffusion models will gradually add noise to $x_0$ with a Markov chain. In the reverse process, a noise-predictor $\epsilon_\theta$ is trained to recover the noisy image $x_t$ by predicting the noise of $t$-step. Then Mean Squared Error (MSE) loss is applied to minimize the distance between the predicted noise $\epsilon_\theta(x_t, t)$ and real noise $\epsilon$. Since SD does not perform the diffusion process in image space but in latent space, we let $z_t$ denote the latent space variable of the \textit{t-step} and ${\varepsilon}_t$ denote the noise added to it at the \textit{t-step} in the diffusion forward process. 

Our goal is to let the noise prediction model (U-Net) predict ${\varepsilon}_t$. 
Therefore, our loss function is designed as followed:
\begin{equation}\label{eqn:our_diffusion_loss}
    \mathcal{L}_\text{\scriptsize \tiny SD} = \mathbb{E}_{z_0, \epsilon, t} \| \epsilon - \epsilon_\theta(z_t, c, t) \|^2,
\end{equation}
where $z_t$ indicates the \textit{t-step} latent vector and $c$ is the condition, i.e., the text-based NFT properties. We fixed the Auto-Encoder (AE) in SD and finetune both the DM's U-Net~\cite{ronneberger2015unet} and CLIP~\cite{radford2021CLIP} text-encoder to bridge the textual and visual gap between the NFT domain and the general image domain. To prevent overfitting, we use a collection-weighted sampling strategy that can reduce the probability of sampling from larger collections. This is important because if uniform sampling is used, SD may overfit NFT images from larger collections while underfitting those from smaller collections. Note that the image generator works together with the Prompt Adaptation module, and plays as a part of actors in RL learning, which can receive rewards from the real market.

\subsection{Multiple Visual Policies for RL Optimization}\label{sec:Reward}
Proper optimization metrics can guide the correct direction of gradient updates, which are one of the crucial factors in improving the quality of the generated NFT images. In this paper, We adopted three visual rewards related to NFT quality for PPO optimization: \textbf{1) Visual market reward}, \textbf{2) Visual aesthetic reward}, and \textbf{3) CLIP cross-modal relevance reward}. We will introduce these reward design details in the following subsections.

\textbf{Visual Market Reward:}
a key to generating more profitable NFT images is mining the visual features related to their market value. Instead of predicting the specific price of an NFT like in previous work, our Market Reward (MR) evaluates the value of NFTs based on their visual features. Our proposed MR consists of a visual feature extraction module followed by a 5-layer Multilayer Perceptron (MLP). Except for the last layer, we adopt LeakyReLU~\cite{xu2015leaklyRELU} to increase non-linearity and BatchNorm~\cite{ioffe2015batch_norm} to stabilize the training process. We fix the visual feature extractor and only train the MLP. The MR is optimized with cross-entropy loss, shown as follows:
\begin{equation}\label{eqn:market_CE_loss}
    \mathcal{L}_\text{\scriptsize \tiny CE}= -\mathbb{E}_{i\in \mathcal{N}} \, (y_{i} \log \hat{y}_{i}),
\end{equation}
where $y_{i}$ equal to $1$ if data $x$ belongs to the $i$-th class, and $0$ otherwise. $\hat{y}_{i}$ is the prediction for data $x$.
The scarcity of high-value NFT images and categories is greatly imbalanced, hence training a market predictor is challenging. To address this issue, we use a category-balanced sampling strategy where each category has an equal probability of being sampled.

We use MR to predict the visual market score, and set the lowest price as reward $0$, and the highest price category as reward $1$. Other categories can be equally divided in the range of [$0$,$1$]. To stabilize PPO training, we measure the improvement in visual market score before and after LLM modification as our market reward. Thus, the final market value reward can be defined as follows:
\begin{equation}\label{eqn:market_reward}
    R_\text{\scriptsize \tiny MKT}= \frac{argmax(\hat{y})}{N_c - 1} - \frac{argmax(\hat{y}^*)}{N_c - 1},
\end{equation}
where the $N_c$ is the number of classes in MR, the $\hat{y}$ and $\hat{y}^*$ represent predicted market value before and after LLM modification, respectively. Although our reward model outputs a discrete value, its accuracy is sufficient to provide a reasonable gradient update direction for subsequent PPO training of the LLM model.

\textbf{Visual Aesthetic Reward:}
aesthetics is another important factor that determines the popularity of an NFT image. To generate more aesthetically pleasing images for NFTs, we apply an aesthetic reward to the final result. We adopt the aesthetic predictor as our aesthetic metrics model, followed by LAION-5B~\cite{schuhmann2022laion_5B}. The aesthetic predictor, consisting of a fixed CLIP visual features extractor and a Multi-Layer Perceptron (MLP), was trained on SAC\footnote{\href{https://github.com/JD-P/simulacra-aesthetic-captions/blob/main/README.md}{simulacra-aesthetic-captions}}, LAION-Logos\footnote{\href{https://laion.ai/}{https://laion.ai/}}, and AVA datasets~\cite{murray2012ava_dataset}  to predict image aesthetic scores.
To make training more stable, we calculate the improvement in aesthetic scores before and after modifying the text with LLM as our reward, which is similar to the mentioned NFT market reward. In addition, as the output of the aesthetic predictor ranges from $1$ to $10$, we further clamp the reward to [-1,1] for normalization. Therefore, our final aesthetic reward is defined as:
\begin{equation}\label{eqn:aesthetic_reward}
    R_\text{\scriptsize \tiny AES} = clamp(F_\text{\scriptsize \tiny AES}(\hat{x}) - F_\text{\scriptsize \tiny AES}(\hat{x}^*), -1, 1),
\end{equation}
where the $clamp(\cdot, a, b)$ is used to restrict $x$ in the interval [a,b], and $F_\text{\scriptsize \tiny AES}(\cdot)$ represents the aesthetic predictor model. $\hat{x}^*$ and $\hat{x}$ represent generated images before and after LLM modification, respectively.

\textbf{CLIP Reward:}
to ensure the semantic consistency between the output image and the user input prompt, we use the CLIP model ~\cite{radford2021CLIP} to calculate the similarity between images and text. Since this aspect is not our main optimization goal, we only apply a penalty when the similarity is below a certain threshold. The objective of this approach is to  encourage the model to prioritize other optimization criteria, i.e., aesthetic rewards and market value rewards. Therefore, our CLIP reward is defined as follows:
\begin{equation}\label{eqn:CLIP_reward}
    R_\text{\scriptsize \tiny CLIP} = \beta_1 * \min(F_\text{\scriptsize \tiny CLIP}(\hat{x}, p) - \zeta, 0),
\end{equation}
where the $F_\text{\scriptsize \tiny CLIP}(\cdot)$ represents the CLIP model, $p$ represents the user input prompt, and $\hat{x}$ is the image generated by our finetuned SD model. $\beta_1$ is used to scale the reward range to [-1, 0]. We empirically set $\beta_1$ as 10 and $\zeta$ as 0.2, which works robustly in practice. It should be noticed that the CLIP reward measures the similarity between generated image and the user input prompt, i.e., the text before LLM modification. Such designs ensure that the generated content can meet the original user intent. 

\begin{figure*}[t]
\setlength{\abovecaptionskip}{-0.05cm} 
\setlength{\belowcaptionskip}{-0.4cm} 
\centering
\includegraphics[width=0.95\textwidth]{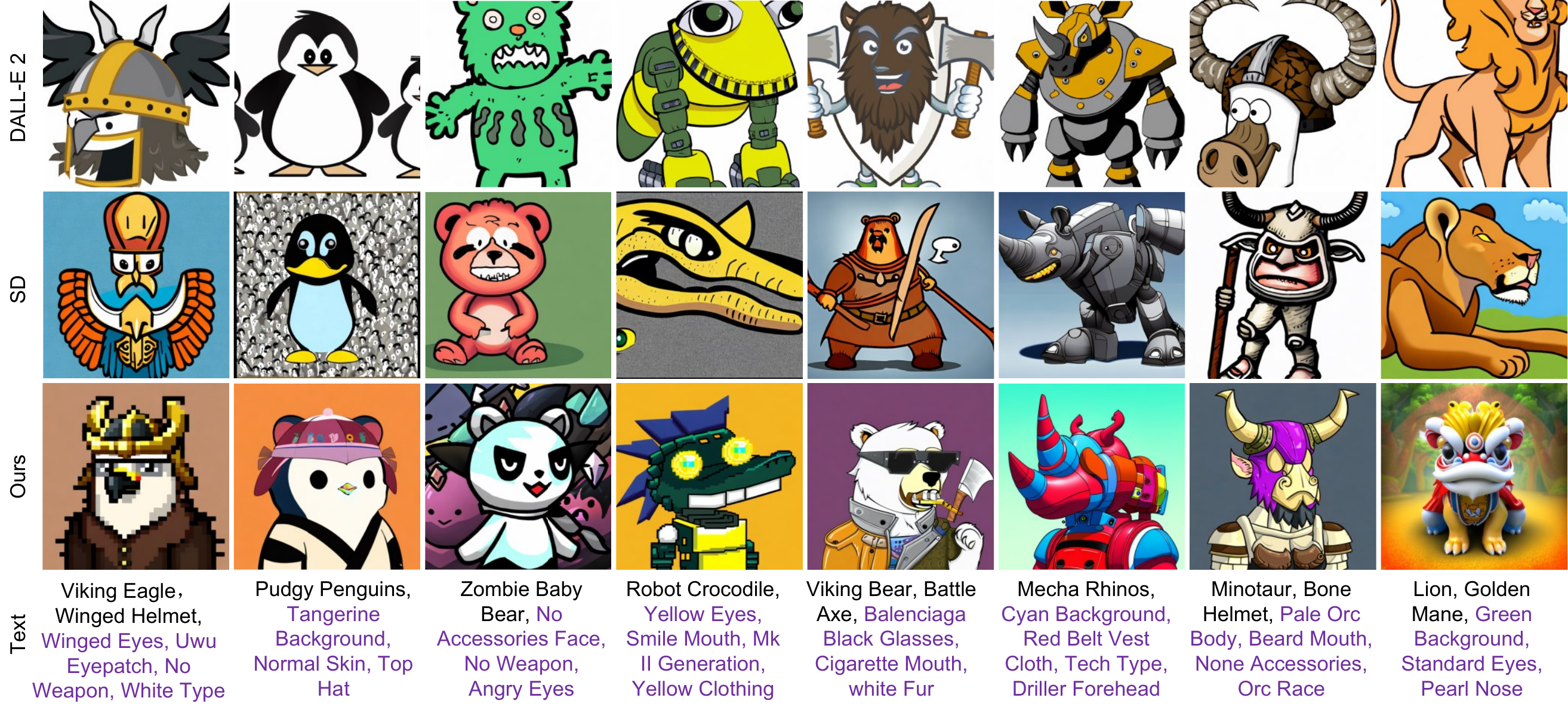}
\caption{Comparisons of text-to-visual NFT generation between \textbf{Diffusion-MVP} and two competitive baseline models, including DALL$\cdot$E 2 and Stable Diffusion (SD). Superior results of our approach can be observed from the more visual-appealing results. The purple texts are completed by our fine-tuned LLM (i.e., GPT-2), given user input objects.}
\label{fig:results}
\end{figure*}

\subsection{Training Strategy}\label{sec:training}

To optimize the proposed framework, we propose a four-step training process for \textbf{Diffusion-MVP}: 1) train the NFT visual market reward model; 2) fine-tune the base NFT image generator (SD); 3) conduct supervised fine-tuning of LLM; 4) train LLM with PPO. Note that before the PPO training, we fine-tune SD and LLM to learn NFT domain knowledge, in the proposed \textbf{NFT-1.5M} dataset. Finally, PPO training can maximize the three visual rewards above to generate more profitable NFT images.

To finetune the LLM, we first randomly shuffle the order of the properties description regarding them as complete outputs. And then, we randomly discard some of them regarding them as inputs. The complete outputs and inputs are combined by a prompt in the format of \textit{"[input][prompt][output]"} as training data. LLM is trained with log-likelihood to maximize the probability of the next token.
We experimentally found that the prompt "\textit{. Add details:}" following user inputs, performed the best in terms of training difficulty and valuation performance. During the PPO training process, we fix SD and train LLM with three visual policies rewards described in Sec.~\ref{sec:market_value}. The total reward is defined as follows:
\begin{equation}\label{eqn:total_loss}
    R = \lambda_1 R_{\text{\scriptsize \tiny MKT}} + \lambda_2 R_\text{\scriptsize \tiny AES} + \lambda_3 R_\text{\scriptsize \tiny CLIP},
\end{equation}
where the $\lambda_1$, $\lambda_2$, and $\lambda_3$ are the weights to different reward terms. Because the market reward is our main optimization target, we empirically set $\lambda_1$ as $1$ and the other two as $0.5$ in our experiments.

\section{Experiments}
In this section, we will introduce implementation details, evaluation metrics, and evaluate the proposed generation framework for NFT images, in terms of both object evaluations and user studies.

\subsection{Implementation Details} \label{sec:implementation_details}

To obtain a base image generator on our NFT dataset, we first fine-tuned SD model for 20k iterations with a batch size of 128 and a learning rate of $10^{-6}$. For the following steps, the batch size is set to 512 and the learning rate is adjusted as $5\times10^{-5}$. All experiments were conducted using the Adam  optimizer~\cite{kingma2014Adam}, which is implemented with the popular framework PyTorch~\cite{paszke2019pytorch}. During the test process, all SD models are sampled in 50 steps using DDIM solver~\cite{song2020DDIM_diffusion}. For PPO training, we follow DPM~\cite{lu2022DPM} solver's 20-step sampling in order to speed up and average the rewards by sampling three images each time to reduce the effect of randomness. More details of the implementation can be found in the supplementary material.

\subsection{Evaluation Metrics} \label{sec:metrics}

We evaluate generated NFT images by using the following four criterias: \textbf{1) resemblance to an NFT image, 2) aesthetics, 3) NFT market value, and 4) consistency between image and text}. To ensure accuracy and reliability, each criteria is evaluated objectively and subjectively, with numerical results reported. 

\textbf{Objective Evaluation:}
we evaluate the accuracy of our market reward model using a non-overlapping test set. The high performance can be found in supplementary materials. Because there is limited research on accurately predicting the value of generated NFT images, we use our own reward model as an objective measure for \textbf{market value prediction}. Inspired by previous works~\cite{sd_ldm_diffusion, ramesh2021zero_T2I_diffusion, nichol2021CLIDE_diffusion, dhariwal2021_diffusion}, we employ the \textbf{Fréchet Inception Distance (FID)}~\cite{heusel2017FID} score to assess the distribution similarity between generated images and NFT images. Additionally, we utilize an aesthetic predictor to determine \textbf{aesthetic scores}.

\textbf{Subjective Evaluation:}
we conducted a user study to assess subjective results. For a fair comparison, we used ChatGPT~\cite{openai2023GPT4} to generate 200 text prompts for NFT image generation, each containing a cartoon character or animal as the main subject and a few descriptive words. All methods generated images based on these prompts for comparison. Due to the instability of individual scores and the wide range of scores among different people, we used a side-by-side comparison in our user study. For each review, we randomly presented two images from different methods along with their corresponding texts. Human subjects chose one of three options for each indicator based on their own judgment: Image A is better, Image B is better, or they are comparable. We invited 10 third-party evaluators (5 male and 5 female subjects) to conduct the evaluation. All of them are familiar with NFT images, and have a uniform distribution over ages (from 20 to 60). Each person reviewed 200 times per round, forming a total of 2k voting scores. Finally, we tallied all the results and presented them as percentages.

\subsection{Comparison with SOTA Methods}

To demonstrate the advantage of our method, we compare our method with the state-of-the-art approaches, DALL$\cdot$E~2~\cite{ramesh2022hierarchical_DALLE2} and SD~\cite{sd_ldm_diffusion}. From both subjective and objective perspectives, we shall compare four indicators: \textbf{1)resemblance to an NFT image, 2) Aesthetics, 3) NFT market value, and 4) consistency between image and text}. 

\textbf{Objective Comparison:}
the overall results can be found in Tab.~\ref{tab:results}. As can be seen from the table, our Diffusion-MVP has surpassed the existing SOTA methods in four metrics, visual Market Value (MV), FID~\cite{heusel2017FID}, Aesthetics score, and CLIP~\cite{radford2021CLIP} similarity.
Specifically, among the metrics MV, aesthetics, and FID, our Diffusion-MVP achieved an improvement of \textbf{0.095 (14.7\%)},  \textbf{0.311 (6.1\%)}, and \textbf{24.82(15.5\%)} compared to DALL$\cdot$E~2, and an improvement of \textbf{0.115 (18.4\%)}, \textbf{0.228 (4.4\%)}, and \textbf{16.41(10.8\%)} compared to SD. Under the CLIP metric, our method slightly outperforms SD and DALL$\cdot$E~2. This small gain is due to our focus on generating valuable NFT images rather than CLIP rewards. In conclusion, all these results effectively prove that our approach can generate more profitable, aesthetic, and NFT-style images while maintaining better semantic consistency, outperforming the existing SOTA methods.

\begin{figure}[t]
    \setlength{\abovecaptionskip}{-0.05cm} 
    \setlength{\belowcaptionskip}{-0.2cm} 
    \centering
    \includegraphics[width=0.95\columnwidth]{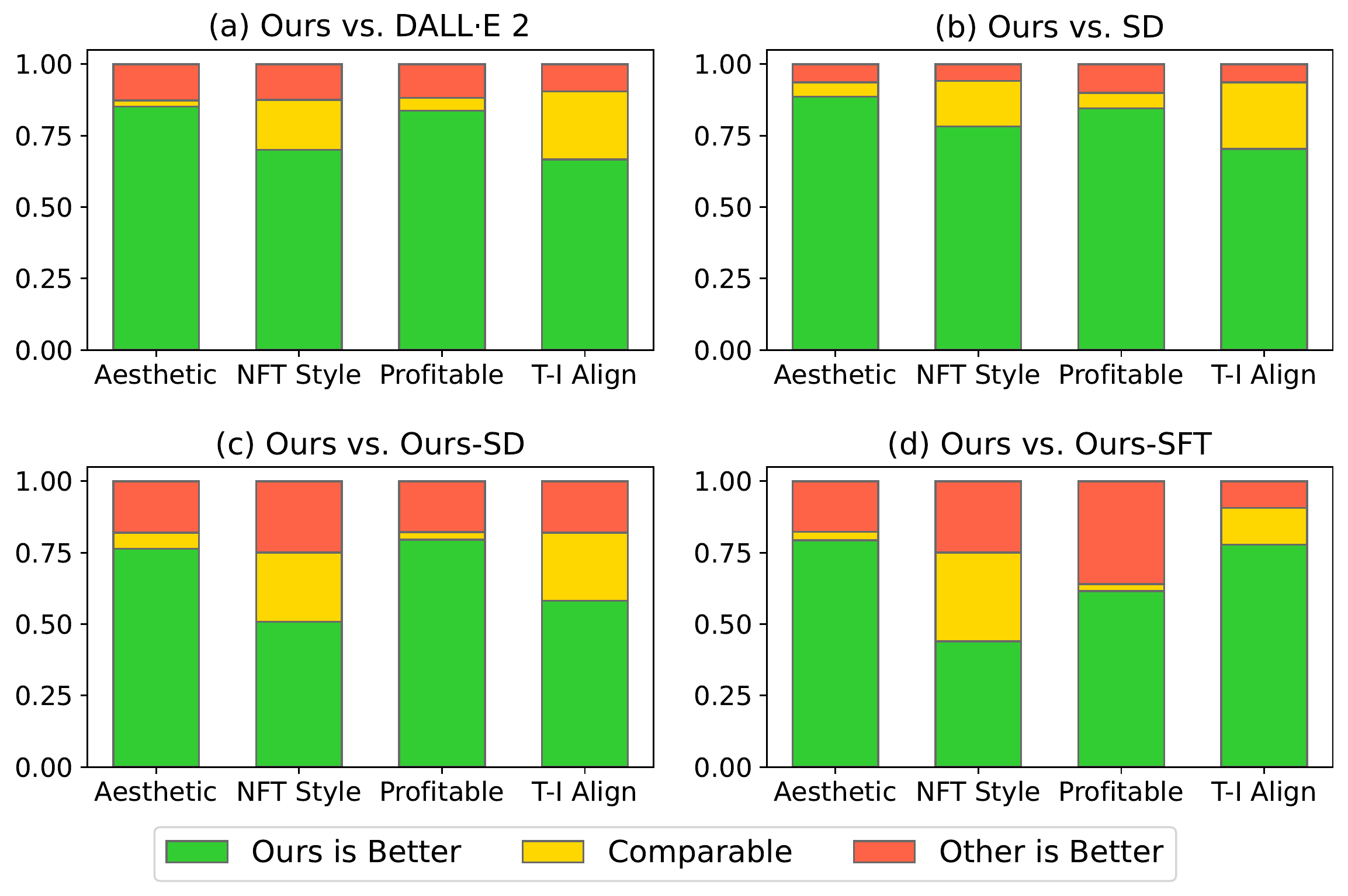}
    \caption{User study conducted by 10 human subjects, on different comparison settings. Dominant preferences to our full model are clearly presented, compared with other competitive baselines (a,b) and ablation models (c,d). T-I Align means text-image relevance.}
    \label{fig:user_study}
\end{figure}

\textbf{Subjective Comparison:}
to prevent the bias of objective metrics, we also conduct the user-studies to further verify our method. The overall subjective comparison results are shown in Fig.~\ref{fig:user_study}. It can be seen from Fig.~\ref{fig:user_study} that over 87\% of evaluators believe that our method surpasses or is comparable to DALL$\cdot$E~2, on four evaluation metrics. Compared to SD, this proportion reaches to 90\%. These results fully demonstrate that the images generated by our method are more visullay-pleasing, more popular, and more profitable, compared with existing SOTA methods.
We also show the generated images of \textbf{Diffusion-MVP} and other SOTA methods in Fig.~\ref{fig:results}. As you can see from Fig.~\ref{fig:results}, the images we generate are more aesthetically pleasing and contain more attractive elements, such as the golden body of the lion (last column), and the bear costume (fifth column). This also proves the effectiveness of our approach.

\begin{table}[]
\setlength{\abovecaptionskip}{-0.02cm} 
\centering
\caption{\label{tab:results}Overall objective comparison of Market Value (MV), Aesthetics score, FID, and text-image CLIP Similarity.}
\begin{tabular}{lcccc}
\toprule
          & \begin{tabular}[c]{@{}c@{}}MV$\uparrow$\end{tabular}
          & \begin{tabular}[c]{@{}c@{}}Aesthetic$\uparrow$ \end{tabular} 
          & FID~\cite{heusel2017FID}$\downarrow$ 
          & CLIP~\cite{radford2021CLIP}$\uparrow$ 
          \\ \bottomrule
SD~\cite{sd_ldm_diffusion}                  & 0.625 & 5.194 & 151.56 &  0.247 \\
DALL$\cdot$E 2~\cite{ramesh2022hierarchical_DALLE2} & 0.645 & 5.111 & 159.97  & 0.249 \\ \hline
Ours-SD                                     & 0.670 & 5.211 & 146.37 &  \textcolor{red}{\textbf{0.250}} \\
Ours-SFT                                    & 0.690 & 5.287 & 137.15 &  0.210 \\
Ours                                        & \textcolor{red}{\textbf{0.740}} & \textcolor{red}{\textbf{5.422}} & \textcolor{red}{\textbf{135.15}} &  \textcolor{red}{\textbf{0.250}} \\ \bottomrule
\end{tabular}
\vspace{-0.4cm}
\end{table}

\subsection{Ablation Studies}
We also conduct ablations to verify the effectiveness for each of our modules. We remove the LLM and the input text, which is directly used to generate an image by our fine-tuned SD (denoted as Ours-SD). We also removed the PPO module. The input text is modified by our SFT’s LLM and then generated by our fine-tuned SD (denoted as Ours-SFT). From both subjective and objective perspectives, we also compare four metrics described in Sec.~\ref{sec:metrics}.

\textbf{Objective Comparison:}
all the objective results are displayed in Tab.~\ref{tab:results}. As can be seen from this table, models gradually improve in terms of MV, aesthetics, and FID metrics after adding each of our modules. After fine-tuning SD on the \textbf{NFT-1.5M} dataset, our SD outperforms the original SD on all three metrics, i.e., MV, aesthetic, and FID. This is gained from our high-quality NFT-1.5M dataset. It should be noted that although Our-SFT also has good results in the first three metrics, it experiences a significant decrease in the CLIP metric, dropping from 0.250 to 0.210. Then, after applying reinforcement learning, the CLIP metric returns to its original level of 0.250 and the other three metrics also gain significant improvements. This result strongly verifies the necessity and effectiveness of our reinforcement learning approach.

\begin{figure}[t]
    \centering
    \includegraphics[width=0.98\columnwidth]{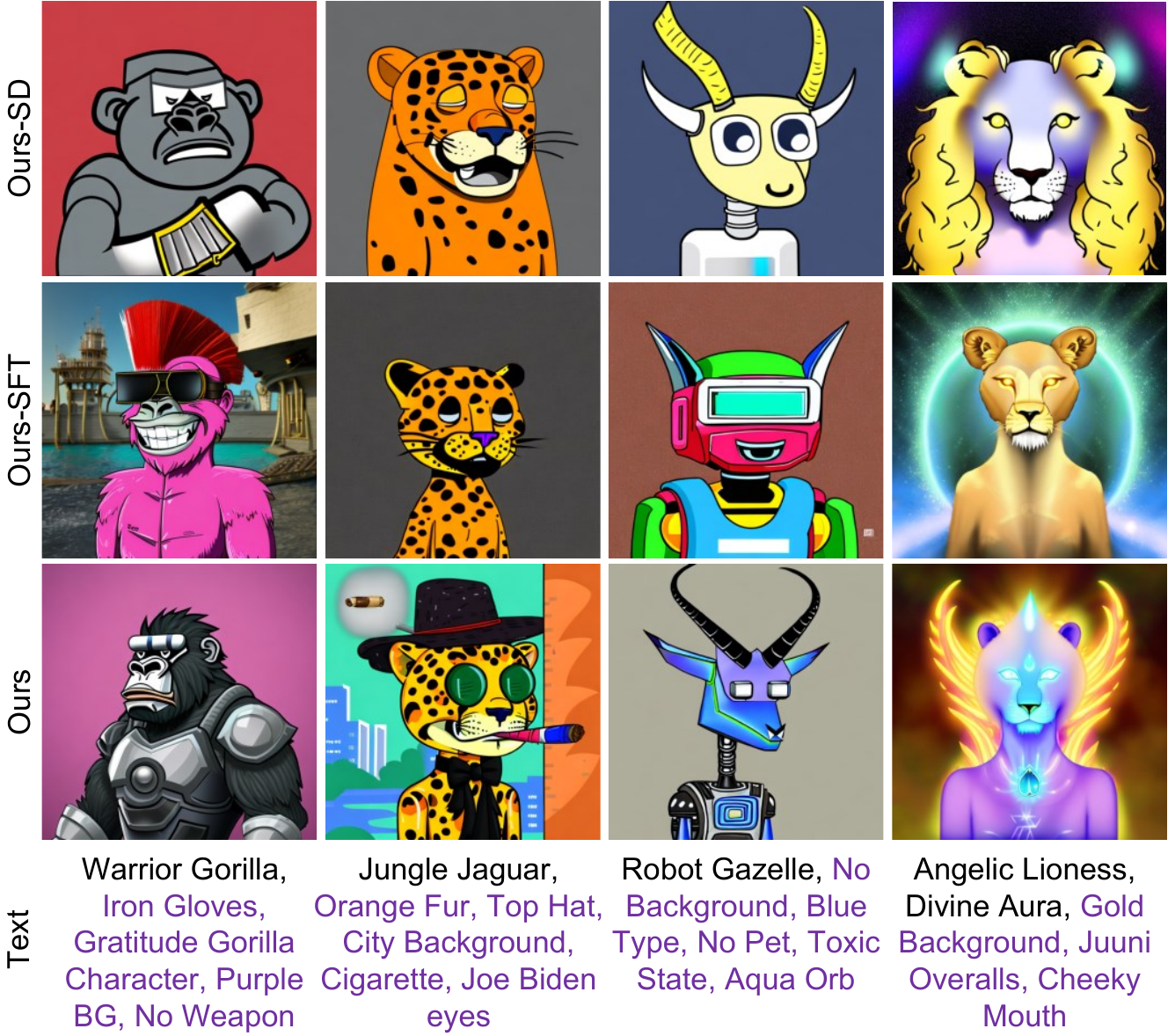}
    \caption{Ablation studies of different generation results.}
    \label{fig:ablfig}
\end{figure}

\textbf{Subjective Comparison:}
the user study result of \textbf{Diffusion-MVP} compared to our different settings, Ours-SD and Ours-SFT, are shown in Fig.~\ref{fig:user_study}.
It can be seen from Fig.~\ref{fig:user_study} that the evaluators prefer \textbf{Diffusion-MVP} to the other settings. In special, after applied reinforcement learning, \textbf{Diffusion-MVP} gains significant improvements over Ours-SFT in all three criteria: aesthetic, profitability, and text-to-image alignment. These results effectively demonstrate the validity of our approach. 
The comparison results of the generated images of \textbf{Diffusion-MVP} compared with Ours-SD and Ours-SFT are shown in Fig.~\ref{fig:ablfig}. It can be observed from Fig.~\ref{fig:ablfig} that the images \textbf{Diffusion-MVP} generate are more visually pleasing and contain plentiful attractive NFT elements.

\section{Conclusion}

In this paper, we have presented a novel Diffusion-based generation framework with Multiple Visual-Policies as rewards (\textbf{Diffusion-MVP}) for generating profitable Non-Fungible Token (NFT) images from user-input texts. Our proposed framework addresses the two key challenges of generating visually-pleasing and highly-profitable NFT images in an automatic way. By incorporating fine-grained visual attribute prompts and effective optimization metrics from NFT markets, our framework is capable of minting NFT images to have both high visual quality and high market value. We have also provided the largest NFT image dataset \textbf{NFT-1.5M} to date. Experimental results demonstrate the effectiveness of our framework in generating NFT images with more visually engaging elements and higher market value, outperforming state-of-the-art approaches. Our work sheds light on the potential of leveraging diffusion models and designing visual policies for generating profitable NFT images, and opens up new avenues for future research in this area.

\begin{acks}
This work is supported by the National Natural Science Foundation of China (61672548, 61173081, U1711262, U1711261, U1811264, U1811261, U1911203, U2001211, U22B2060), Guangdong Basic and Applied Basic Research Foundation (2019B1515130001), Key-Area Research and Development Program of Guangdong Province ( 2020B0101100001).
\end{acks}

\bibliographystyle{ACM-Reference-Format}
\bibliography{ref}

\clearpage
\appendix
\input{supp.tex}

\end{document}

%% file: supp.tex
\section*{Supplementary Material}

This supplementary material provides a comprehensive introduction to Non-Fungible Tokens (NFTs) in Sec.~\ref{sec:NFT}, followed by an extensive review of related works on NFT value evaluation in Sec.~\ref{sec:related_work_NFT}. Detailed information regarding the implementation of our methods is presented in Sec.~\ref{sec:more_implements_details}. In Sec.~\ref{sec:MV_accuracy}, we demonstrate the accuracy of our Market Value (MV) predictor through the results. Finally, additional results are presented in Sec.~\ref{sec:more_results}.

\let\svthefootnote\thefootnote
\let\thefootnote\relax\footnote{\textsuperscript{*}Huan Yang, Jianlong Fu and Jian Yin are the corresponding authors.}
\let\thefootnote\svthefootnote

\section{NFT}\label{sec:NFT}
Web3 refers to a decentralized internet owned by its builders and users and orchestrated through the use of tokens. Within this ecosystem, tokens represent value or utility and can be classified into two distinct categories: fungible and non-fungible (NFTs). Fungible tokens, such as Bitcoin~\footnote{\href{https://bitcoin.org/en/}{bitcoin.org}}, ETH~\footnote{\href{https://ethereum.org/en/}{ethereum.org}}, and Dogecoin~\footnote{\href{https://dogecoin.com/}{dogecoin.com}}, are interchangeable and can be likened to traditional currencies or stocks. In contrast, NFTs are usually digital artwork, such as a painting or a video, which are unique and non-interchangeable. 

NFT is a type of digital certificate built on blockchain technology, e.g., Ethereum, that guarantees ownership of a unique digital asset. Minting digital assets, such as art, music, or articles, as NFTs, is one way for artists to monetize their work. The other more innovative use for NFTs is the ability to guarantee credit for the original creation. Since NFTs are recorded on a blockchain, the creator of the NFT is recorded in the public ledger. This record in the ledger allows the creator to set a fee, known as a royalty, for whenever the digital asset is sold in the future and earn passive income over time if their work is sold on the secondary market. As a result, more and more artists and users are creating, trading, and collecting these NFT assets in the NFT Marketplace.

Popular NFTs are often released in the form of collections, where each NFT within a collection shares similar characters or themes.  Among the top 1000 most popular NFT collections, approximately 90\% are released in the form of images. Within the same collection, different NFTs may vary in their finer details. The visual feature, such as the richness and rarity of the elements, affect their popularity in the market and, consequently, their price. This phenomenon has also been revealed in previous works~\cite{mekacher2022SCI}.
Fig.~\ref{fig:NFT_display} in the main paper shows examples of a popular NFT collection and its corresponding prices. It can be observed that the prices of NFTs are influenced by the richness and attractiveness of their character attributes.  For instance, as shown in Fig.~\ref{fig:NFT_display} in the main paper, the price of \textbf{BEANZ} increases as its clothing becomes more luxurious and attractive.

In essence, the visual characteristics of an NFT can influence its market value. However, the task of identifying and extracting these valuable visual features to generate more profitable NFT images is far from trivial. This is also the objective of this paper.

\begin{table}[]
\centering
\caption{\label{tab:marketpre}Market value prediction by our designed visual policy model. The principal diagonal shows the correct prediction.}
\begin{tabular}{cc|ccc}
\hline
\multicolumn{2}{c|}{\multirow{2}{*}{\begin{tabular}[c]{@{}c@{}}Confusion\\ Matrix\end{tabular}}} & \multicolumn{3}{c}{Ground Truth} \\ \cline{3-5} 
\multicolumn{2}{c|}{} & Low & Medium & High \\ \hline
\multicolumn{1}{c|}{\multirow{3}{*}{Prediction}} & Low & \textbf{33.0\%} & 2.8\% & 1.1\% \\
\multicolumn{1}{c|}{} & Medium & 2.3\% & \textbf{24.5\%} & 3.9\% \\
\multicolumn{1}{c|}{} & High & 0.4\% & 3.9\% & \textbf{28.1\%} \\ \hline
\end{tabular}
\end{table}

\section{Related Works of NFT Value Evaluation} \label{sec:related_work_NFT}

NFT is in its infancy and several works~\cite{NFT_Brand, horky2022price, tweetboost, mekacher2022SCI, NFTmapping, NFTWWW2022char, NFTWWW2023show} have attempted to evaluate NFT Value. For example, Colicev et al.~\cite{NFT_Brand} revealed that NFTs can bring value to brands by representing brand components, attracting brand awareness, generating cross-selling opportunities, and forming highly engaging brand communities. Horky et al.~\cite{horky2022price} attempted to use quantitative tools to predict the value of NFTs and showed that NFTs cannot be simply regarded as an extension of cryptocurrencies. Several studies have shown that the price of NFTs is influenced by social information~\cite{tweetboost}, their rarity~\cite{mekacher2022SCI}, and their multi-modal feature~\cite{NFTmapping, NFTWWW2022char, NFTWWW2023show}. Specifically, Nadini et al.~\cite{NFTmapping} found that historical sales prices and visual clustering features are good predictors of NFT prices. Costa et al.~\cite{NFTWWW2023show} also found that using visual and text information can predict the price range of NFTs with pleasing results. These studies have demonstrated that  the presence of rare visual features in NFT contributes to their higher market value. However, there is a lack of research investigating the integration of market value into the NFT generation process. This paper aims to bridge that gap by exploring the incorporation of rarity-orient market value into the creation of NFTs. By doing so, this study provides valuable insights into NFT creators, enabling them to generate profitable NFTs that capitalize on rarity-orient market value.

\section{Experiments}
In this section, we first introduce more implementation details in Sec.~\ref{sec:more_implements_details}. Then we show the accuracy of our  Market Value (MV) predictor in Sec.~\ref{sec:MV_accuracy}. Finally, additional results can be found in Sec.~\ref{sec:more_results}.

\subsection{More Implementation Details} \label{sec:more_implements_details}
 
Sec.~5.1 of the main paper presents basic implementation details. We provide additional details here. The Stable Diffusion (SD) is initialized with the parameters of SDv2-1-base~\footnote{\href{https://huggingface.co/stabilityai/stable-diffusion-2-1-base}{stable-diffusion-2-1-base}} and finetuned on 32 NVIDIA V100-32G GPUs for approximately four days using half-precision to accelerate training and reduce memory consumption. When training \textbf{Diffusion-MVP} with Proximal Policy Optimization (PPO), we adopt a Kullback-Leibler (KL) penalty to prevent the LLM model from deviating significantly from SFT. The weight of this KL-penalty is set to 0.2, while the weights of policy gradient loss ($\mathcal{L}_\text{\scriptsize \tiny PG}$ in Eqn.~3) and critic loss ($\mathcal{L}_\text{\scriptsize \tiny V}$ in Eqn.~5) are set to 1 and 0.2, respectively.

\subsection{Accuracy of Market Value} \label{sec:MV_accuracy}
The Market Value (MV) predictor is crucial for mining value-related information in our \textbf{Diffusion-MVP} due to an accurate MV predictor providing a good gradient descent for PPO to improve the training effectiveness. 

To verify the effectiveness of our MV predictor, we randomly divided a non-overlapping test set from the \textbf{NFT-1.5M} dataset, comprising about 2k images with an equal number of samples from each category.
The overall accuracy of model predictions was 85.62\% and the detailed confusion matrix is displayed in Tab.~\ref{tab:marketpre}. We can observe from Tab.~\ref{tab:marketpre} that the MR has comparable accuracy in different categories and can distinguish well between low-priced and high-priced categories. This also guarantees that the correct gradient descent during PPO optimization is accurate.

\subsection{More Generation Results} \label{sec:more_results}
\begin{figure*}[t]
\centering
\includegraphics[width=0.95\textwidth]{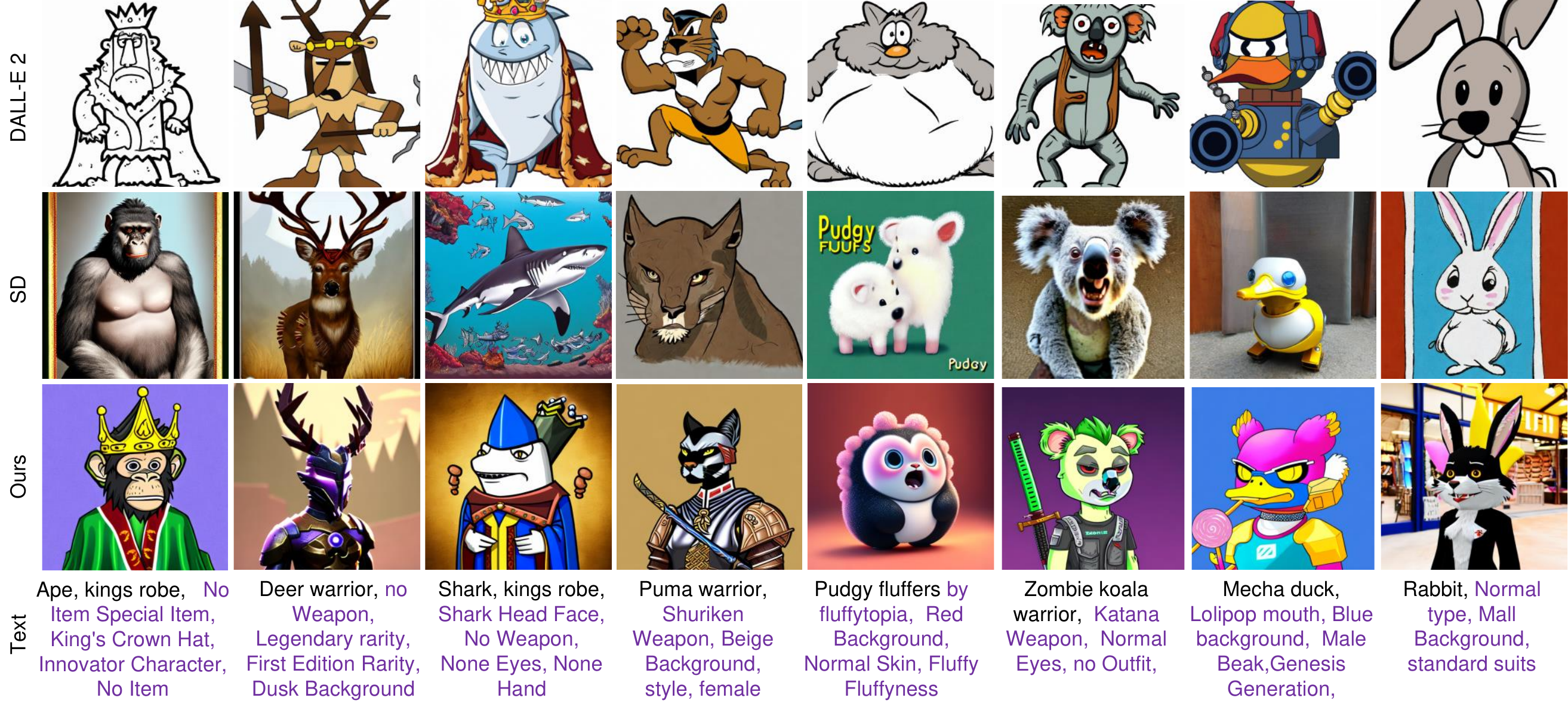}
\caption{Comparisons of text-to-visual NFT generation between \textbf{Diffusion-MVP} and two competitive baseline models, including DALL$\cdot$E 2 and Stable Diffusion (SD). Superior results of our approach can be observed from the more visual-appealing results. The purple texts are completed by our fine-tuned LLM (i.e., GPT-2), given user input objects.}
\label{fig:main_fig_supp1}
\end{figure*}

\begin{figure*}[t]
\centering
\includegraphics[width=0.95\textwidth]{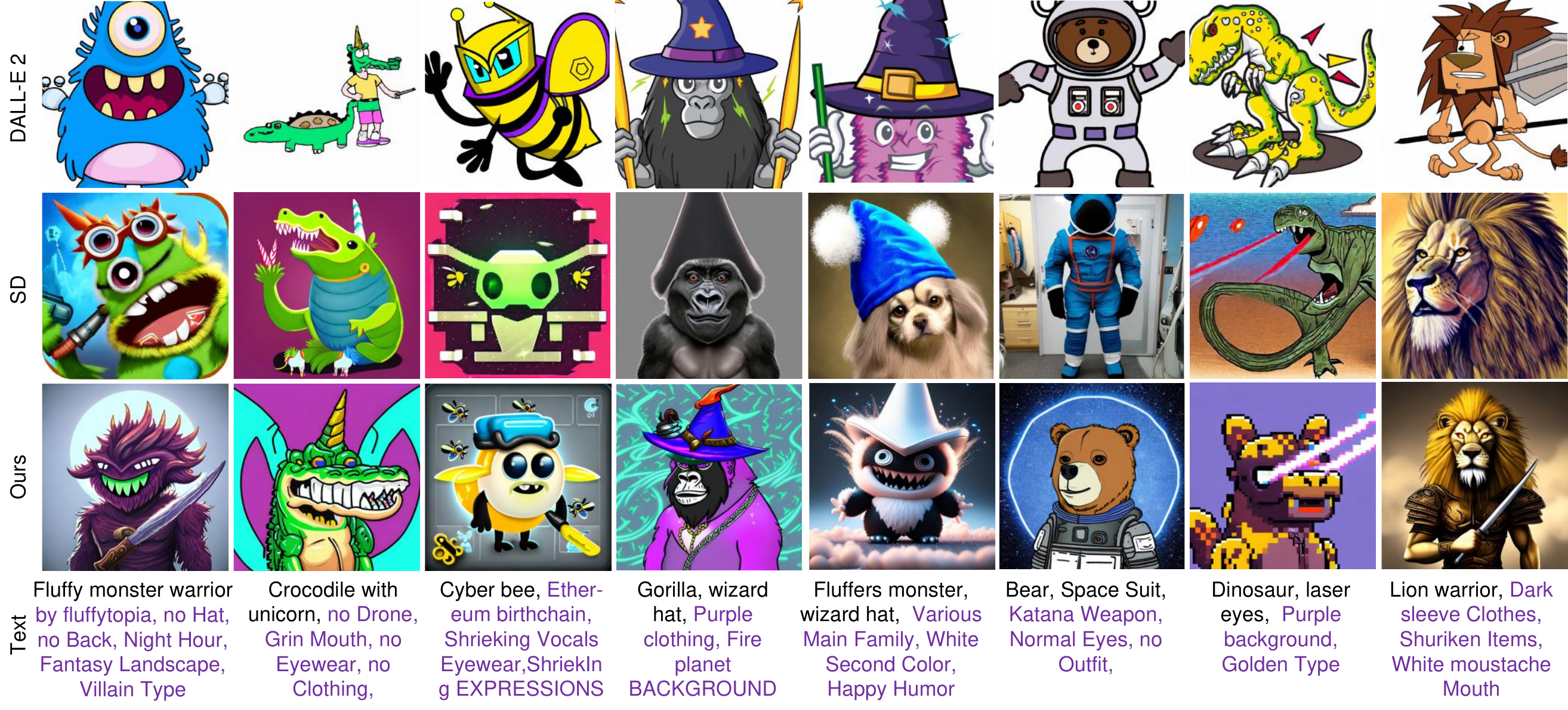}
\caption{Comparisons of text-to-visual NFT generation between \textbf{Diffusion-MVP} and two competitive baseline models, including DALL$\cdot$E 2 and Stable Diffusion (SD). Superior results of our approach can be observed from the more visual-appealing results. The purple texts are completed by our fine-tuned LLM (i.e., GPT-2), given user input objects.}
\label{fig:main_fig_supp2}
\end{figure*}

\begin{figure*}[t]
\centering
\includegraphics[width=0.95\textwidth]{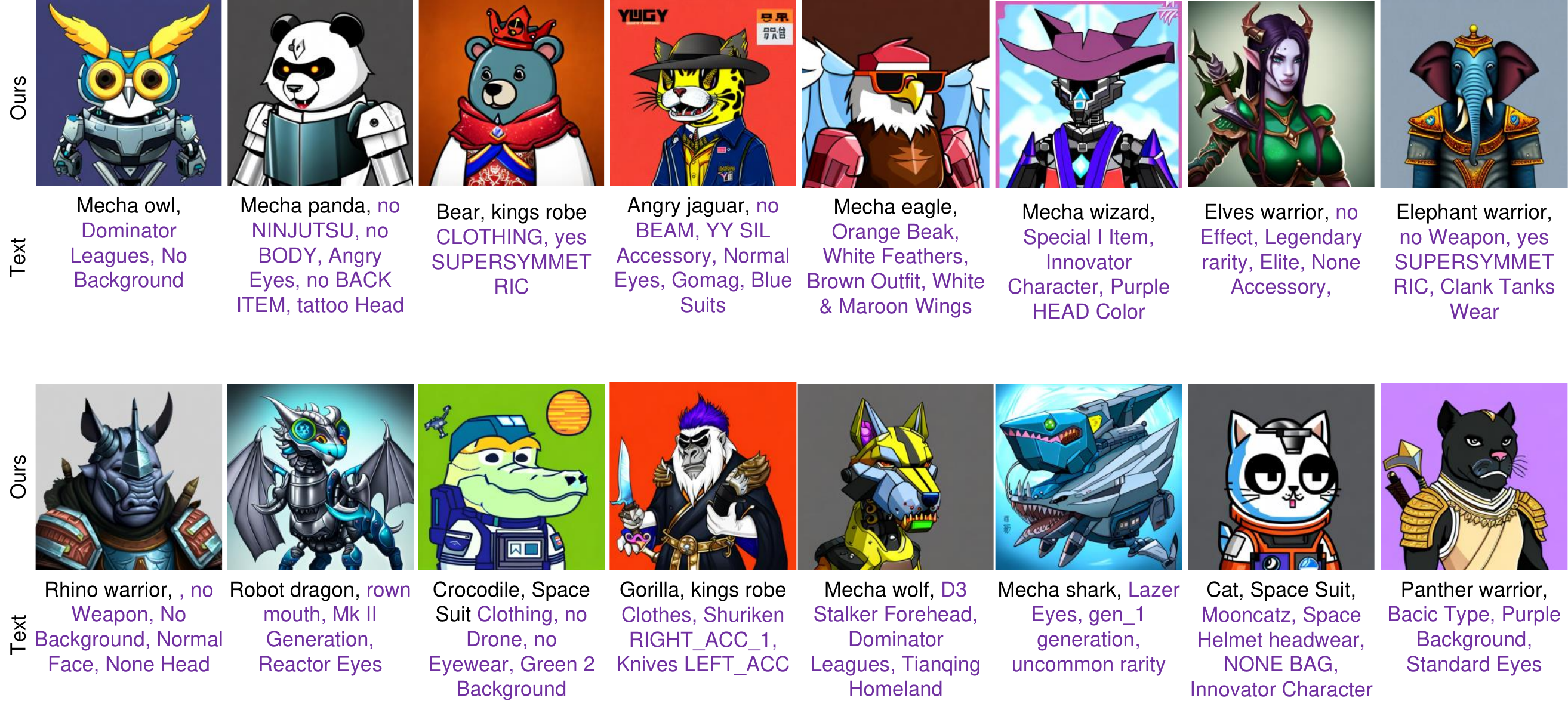}
\caption{Generated images from our \textbf{Diffusion-MVP} approach. Our approach produces visually appealing results, demonstrating its superiority. The purple texts are completed by our fine-tuned LLM (i.e., GPT-2), given user input objects.}
\label{fig:ours_supp1}
\end{figure*}

\begin{figure*}[t]
\centering
\includegraphics[width=0.95\textwidth]{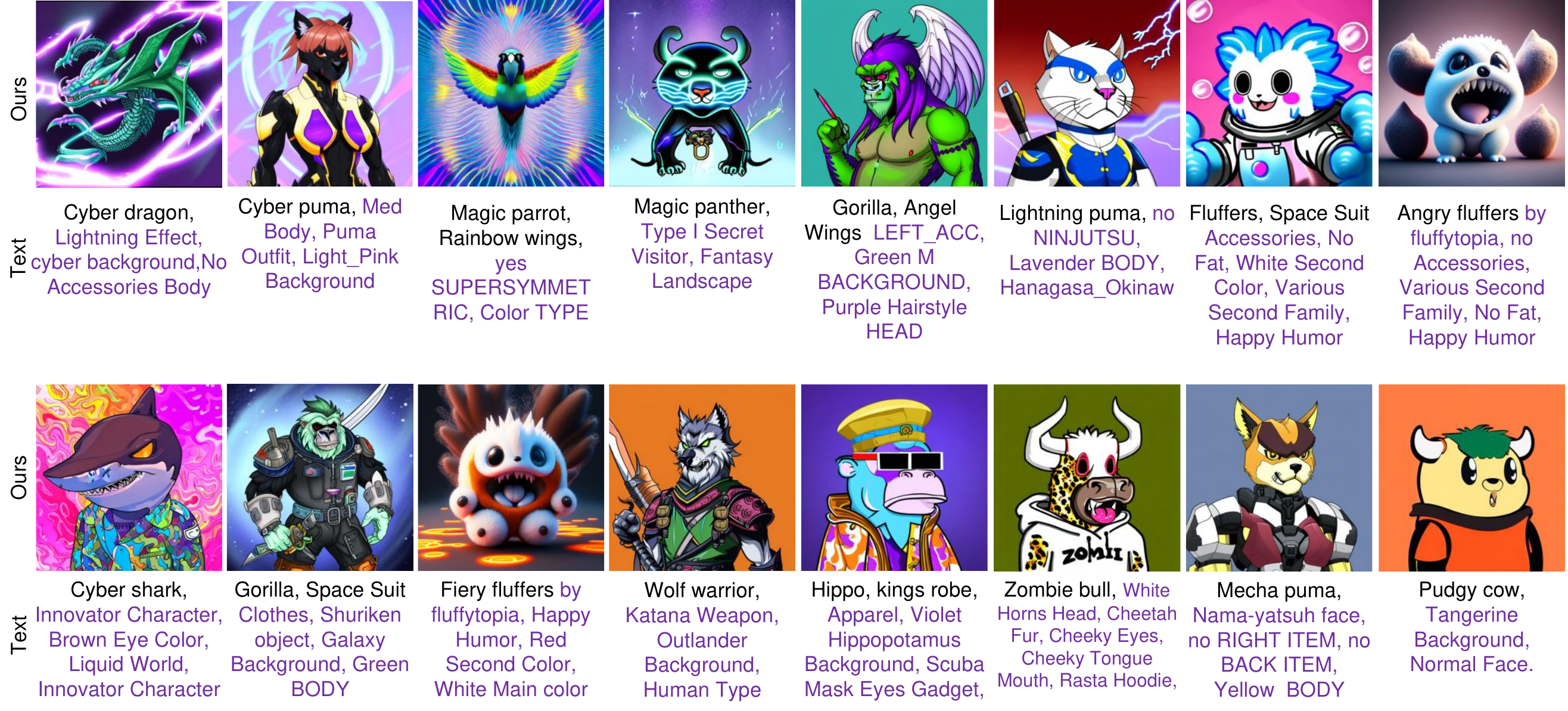}
\caption{Generated images from our \textbf{Diffusion-MVP} approach. Our approach produces visually appealing results, demonstrating its superiority. The purple texts are completed by our fine-tuned LLM (i.e., GPT-2), given user input objects.}
\label{fig:ours_supp2}
\end{figure*}

In Fig.~\ref{fig:results} and Fig.~\ref{fig:ablfig} of the main paper, we presented a selection of our generated images. In this subsection, we provide additional results in Fig.~\ref{fig:main_fig_supp1}, Fig.~\ref{fig:main_fig_supp2}, Fig.~\ref{fig:ours_supp1}, and Fig.~\ref{fig:ours_supp2}. As can be seen from Fig.~\ref{fig:main_fig_supp1} and Fig.~\ref{fig:main_fig_supp2}, \textbf{Diffusion-MVP} generates images that are more NFT-style and have richer and attractive elements compared to existing SOTA methods SD~\cite{sd_ldm_diffusion} and DALL$\cdot$E~2~\cite{ramesh2022hierarchical_DALLE2}. This fully verifies the effectiveness of our method.

\begin{figure*}[t]
\centering
\includegraphics[width=0.95\textwidth]{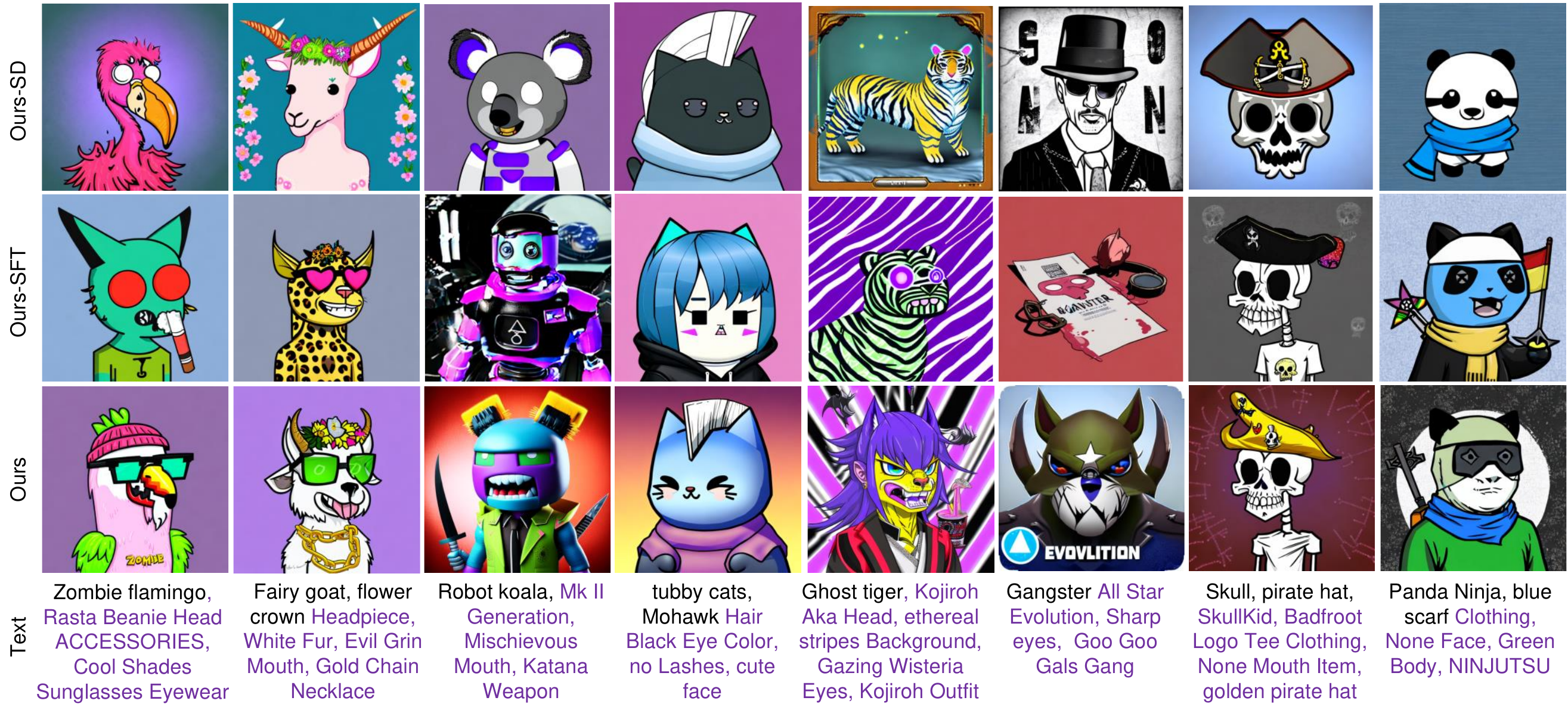}
\caption{Ablation studies of different generation results. Superior results of our approach can be observed from the more visual-appealing results. The purple texts are completed by our PPO-trained LLM (i.e., GPT-2), given user input objects.}
\label{fig:abl-supp1}
\end{figure*}

\begin{figure*}[t]
\centering
\includegraphics[width=0.95\textwidth]{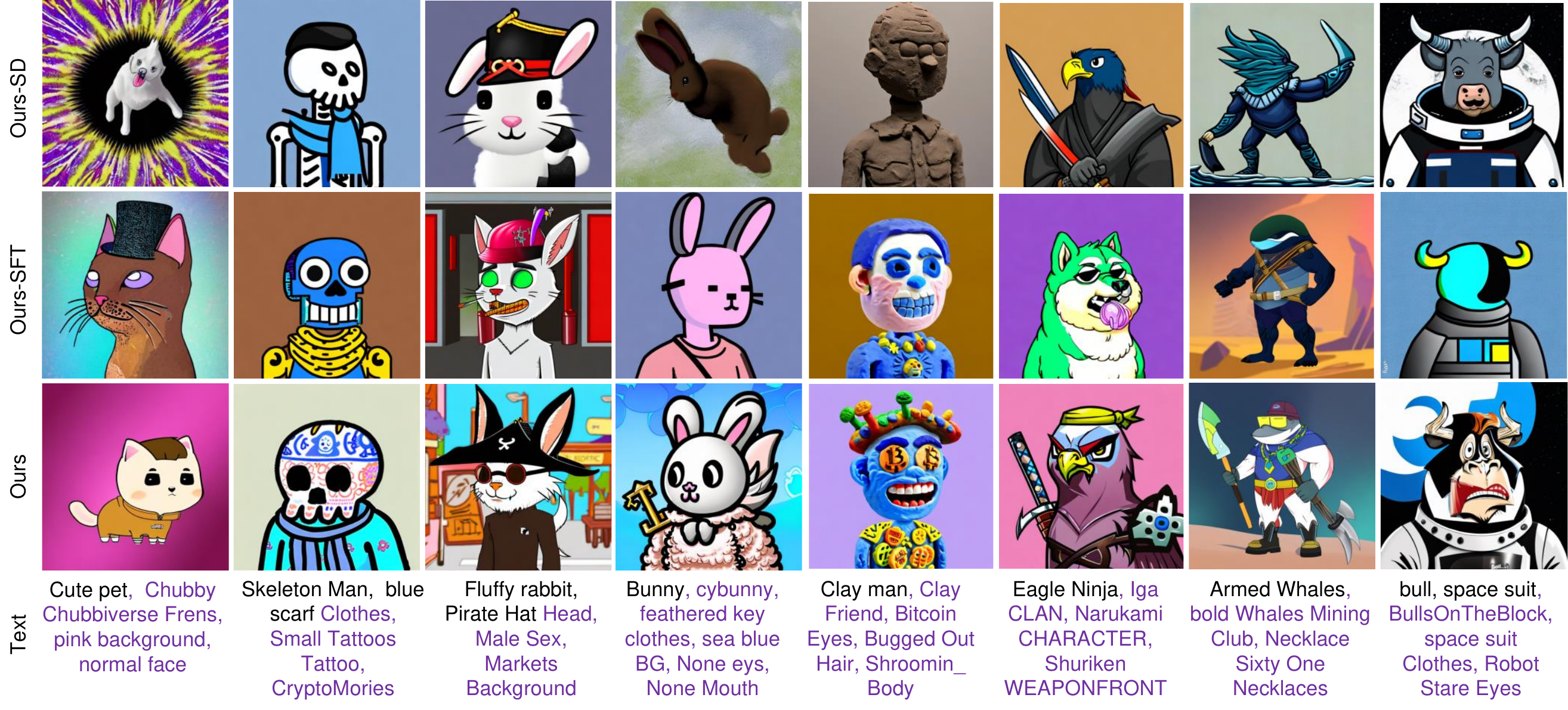}
\caption{Ablation studies of different generation results. Superior results of our approach can be observed from the more visual-appealing results. The purple texts are completed by our PPO-trained LLM (i.e., GPT-2), given user input objects.}
\label{fig:abl-supp2}
\end{figure*}

We also present more visual results of our ablation study in Fig.~\ref{fig:abl-supp1} and Fig.~\ref{fig:abl-supp2}. As shown in these figures, our approach, augmented with reinforcement learning, produces images that encompass more aesthetically pleasing and distinctive elements. For instance, the golden pirate hat on the skull in column 7 of Fig.~\ref{fig:abl-supp1} and the bitcoin eyes and mushroom body of the clay man in column 5 of Fig.~\ref{fig:abl-supp2} are rarer and more visually appealing. Consequently, our method has achieved higher market value. This fully confirms the effectiveness of our approach.

